\documentclass{article}
\usepackage{spconf}
\usepackage{amsmath,graphicx}
\usepackage{cite}
\usepackage{amssymb}
\usepackage{amsfonts}
\usepackage{algorithmic}
\usepackage{algorithm}
\usepackage{array}
\usepackage{subfigure}
\usepackage{textcomp}
\usepackage{stfloats}
\usepackage{url}
\usepackage{verbatim}
\usepackage{bm}
\usepackage{threeparttable}
\usepackage{multirow}
\usepackage{color}
\usepackage[table,xcdraw]{xcolor}

\title{Level-line Guided Edge Drawing for Robust Line Segment Detection}

\name{Xinyu Lin$^{1}$ \qquad Yingjie Zhou$^{2}$ \qquad Yipeng Liu$^{1}$ \qquad Ce Zhu$^{1*}$ \thanks{* Corresponding author. This research is supported by the National Natural Science Foundation of China (NSFC, No. U19A2052, 62171088, and 62171302), and in part by the Sichuan Youth Science and Technology Innovation Team under Grant 2022JDTD0014.}}

\address{$^{1}$School of Information and Communication Engineering, \\University of Electronic Science and Technology of China \\ $^{2}$College of Computer Science, Sichuan University}

\begin{document}
    %\ninept	
	
	\maketitle
	\begin{abstract}
		Line segment detection plays a cornerstone role in computer vision tasks. Among numerous detection methods that have been recently proposed, the ones based on edge drawing attract increasing attention owing to their excellent detection efficiency. However, the existing methods are not robust enough due to the inadequate usage of image gradients for edge drawing and line segment fitting. Based on the observation that the line segments should locate on the edge points with both consistent coordinates and level-line information, \textit{i.e.}, the unit vector perpendicular to the gradient orientation, this paper proposes a level-line guided edge drawing for robust line segment detection (GEDRLSD). The level-line information provides potential directions for edge tracking, which could be served as a guideline for accurate edge drawing. Additionally, the level-line information is fused in line segment fitting to improve the robustness. Numerical experiments show the superiority of the proposed GEDRLSD\footnote{https://github.com/roylin1229/GEDRLSD} algorithm compared with state-of-the-art methods.
	\end{abstract}


	\section{Introduction}
	\label{sec_intro}
	
	Like point features \cite{4217309}, line segments are also the basic image local features for many computer vision tasks, \textit{e.g.}, indoor frame recovery \cite{6638003}, in which line segments are detected and matched to provide the feature correspondences between different images of the same scenes. Line segment detection should be robust and fast to accomplish these tasks. Generally, line segments appear in the image areas where the gradients have a trending change, \textit{e.g.,} the edges \cite{6853580}. 
	
	A lot of line segment detection methods have been proposed recently. According to their detection mechanism, they can be classified into the following three groups roughly: (1) Hough based methods \cite{RobustDetectionofLinesUsingtheProgressiveProbabilisticHoughTransform,MCMLSD}; (2) Local information analysis based methods \cite{LSDaLineSegmentDetector,EDLines,ELSED,Outdoorplacerecognitioninurbanenvironmentsusingstraightlines,ANovelLineletBasedRepresentationforLineSegmentDetection}; (3) Deep learning based methods \cite{TowardsRealtimeandLightweightLineSegmentDetection,End-to-EndWireframeParsing,DeepHoughTransformLinePriors,FullyConvolutionalLineParsing}. Among these methods, the local information analysis based ones attract increasing attention for their high detection efficiency, which is critical for embedded devices with limited computation and storage resources. Notably, the methods \cite{EDLines,ELSED} based on edge drawing \cite{TOPAL2012862} are ultra-fast, which are more than ten times faster than the benchmark real-time line segment detection algorithm \cite{LSDaLineSegmentDetector}. However, they are not robust enough due to inadequate usage of image gradients.
	
	\begin{figure}[tbp]
		\centering
		\setlength{\abovecaptionskip}{0.cm}
		\setlength{\belowcaptionskip}{-0.cm}
		\includegraphics[width=0.91\linewidth]{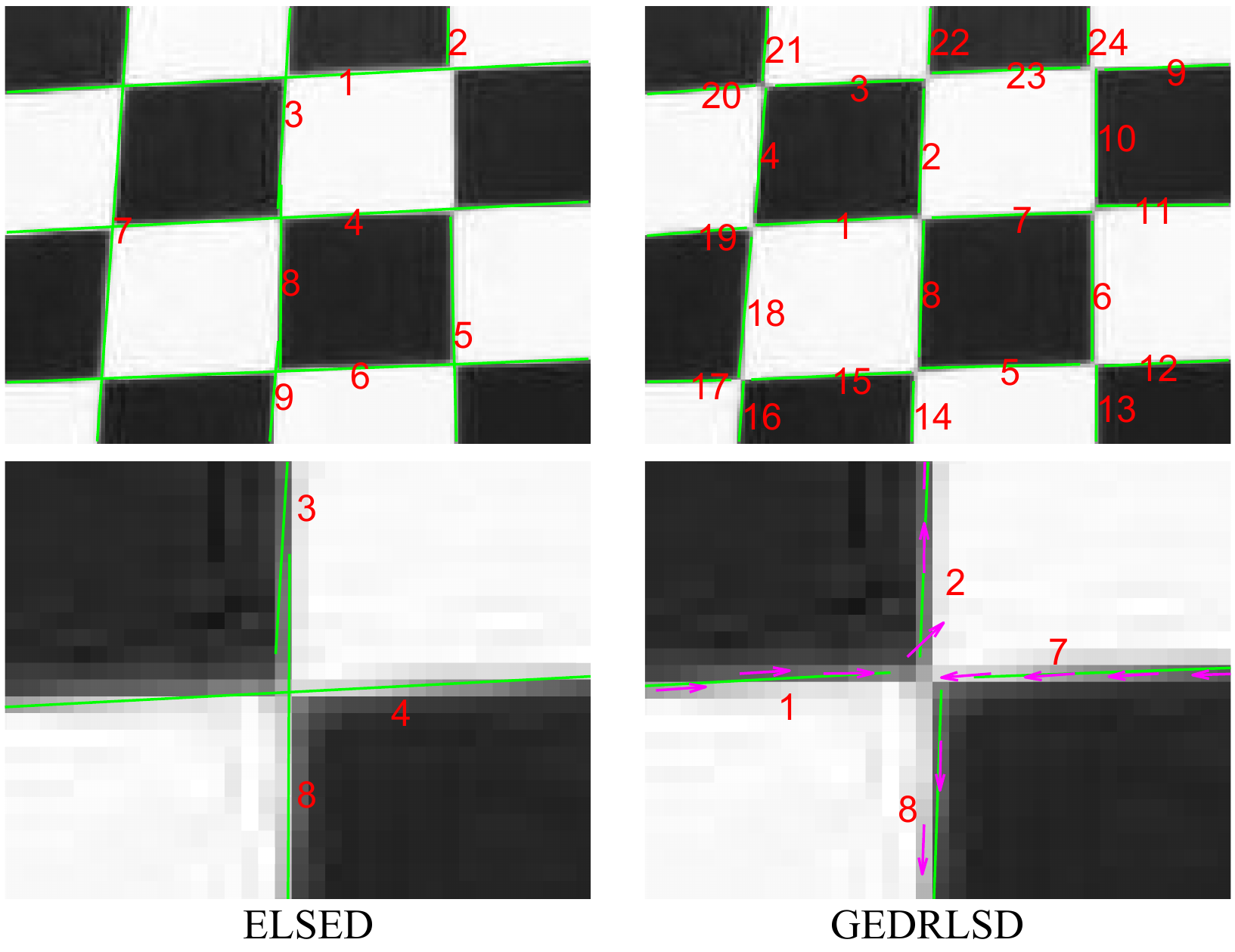}
		\caption{Line segments with details detected by the ELSED \cite{ELSED} and the proposed GEDRLSD methods. The line segments in GEDRLSD should have consistent edge point coordinates and level-lines (magenta arrows), \textit{e.g.}, line 1 and 7 are two lines instead of one because their level-lines are entirely different.}
		\label{fig_line_demo}
	\end{figure}	

\begin{figure}[tbp]
		\centering
		\setlength{\abovecaptionskip}{0.cm}
		\setlength{\belowcaptionskip}{-0.cm}
		\includegraphics[width=1\linewidth]{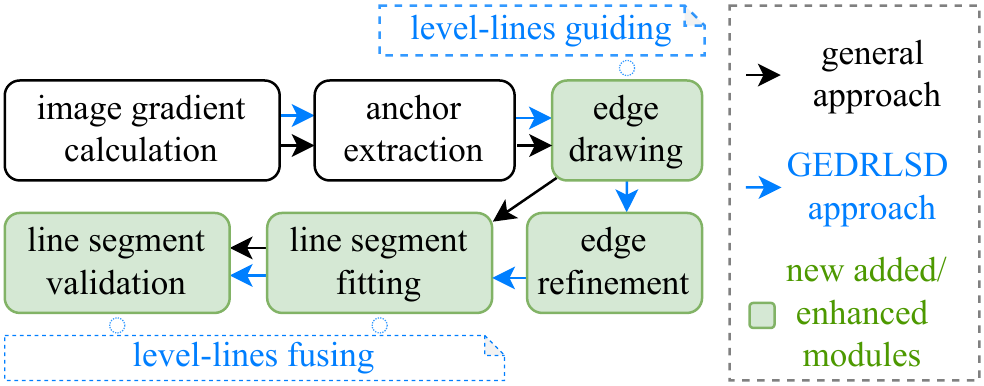}
		\caption{The flowchart of the general edge drawing based approach \textit{vs.} the proposed GEDRLSD approach.}
		\label{fig_flowchart}
	\end{figure}

	As shown in Fig. \ref{fig_line_demo} and \ref{fig_flowchart}, this paper proposes a level-line guided edge drawing for robust line segment detection (GEDRLSD) algorithm by observing that line segments should locate on edge points with both consistent coordinates and level-lines. The double consistent constraints make it more robust than the methods based on a single coordinate constraint, \textit{e.g.}, \cite{EDLines} and \cite{ELSED}. The level-lines are perpendicular to the gradient orientation \cite{LSDaLineSegmentDetector}, which can assign more accurate tracking directions for edge drawing than that in \cite{EDLines} and \cite{ELSED}, significantly when the edge directions are changed, as shown in Fig. \ref{fig_ed} and \ref{fig_edges}. The drawn edges are further refined by analyzing their geometric characteristics, \textit{i.e.}, the reordering for "loop" edges and merging for "line" edges shown in subsection \ref{sec_III_2}. Both edge coordinates and level-lines are fused in line segment fitting under the optimization framework, as shown in formula \ref{loss}. Experiments show that the GEDRLSD method outperforms other state-of-the-art (SOTA) methods while still keeping its efficiency competitive. 
	
	The rest of this paper is organized as follows. In Section \ref{sec_II}, the pipeline of edge drawing based line segment detection algorithms will be described briefly. The proposed GEDRLSD algorithm will be introduced in Section \ref{sec_III}. Numerical experiments based on well-known benchmark datasets are included in Section \ref{sec_IV}. Section \ref{sec_V} is the conclusion of this paper.
	
	\section{Edge Drawing Based Line Segment Detection}
	\label{sec_II}
	Fig. \ref{fig_flowchart} shows the flowchart of general edge drawing based line segment detection. The edges are drawn by the smart routing strategies \cite{EDLines,ELSED}, which connect a series of anchors according to the gradient map. The anchors have a high probability of being edge points, which are pixels with the local maximum of image gradients. When calculating the gradient maps, the original images are generally smoothed by Gaussian filters to decrease the negative effect of noises.
	
	Based on the drawn edge chain, the line segments are fitted progressively in the way of least squares using the coordinates of edge points until the predefined conditions are satisfied, \textit{e.g.}, the distance threshold in \cite{EDLines}. Finally, the fitted line segments can be further validated optionally based on additional constraints, \textit{e.g.}, the constraint of gradient orientation or statistics like the Helmholtz principle in \cite{EDLines}.

	\section{The Proposed GEDRLSD Algorithm}
	\label{sec_III}
	In this paper, except for image gradient magnitude, the level-line information is fully used in the overall process. As shown in Fig. \ref{fig_flowchart}, in contrast to the pipeline shown in Section \ref{sec_II}, the main innovations lie in the \textit{level-line guided edge drawing}, \textit{edge refinement (new process)}, and \textit{level-line leveraged line segment fitting (simultaneously including validation)}.
	
	\subsection{Level-line Guided Edge Drawing}
	\label{sec_III_1}
	Inspired by the general processes for edge drawing, the Gaussian filter with a kernel size of $5\times5$ (w.r.t.  $\delta=1$) is applied in the original image to reduce the noises, and the Sobel filter is used to calculate the image gradients. These pixels with normalized gradient magnitude smaller than a threshold $T^m_{g}$ are not considered in edge drawing. The anchors are extracted in images where the pixels have local maxima of gradient magnitude for each quantized direction (described below) and equalized with a radius of 10 pixels as in \cite{ORB-SLAM} to reduce their numbers and improve the efficiency of edge drawing.

	The level-lines of pixels defined in \cite{LSDaLineSegmentDetector} are perpendicular to their gradient orientation, which can be formulated as 
	\begin{equation}
		\small
		\setlength{\abovecaptionskip}{3.cm}
		\setlength{\belowcaptionskip}{-3.cm}
		\left\{
		\begin{array}{lr}
			{u}       = cos({\theta_o}+\pi /2), &  \\
			{v}       = -sin({\theta_o}+\pi / 2), 
		\end{array}
		\right.
	\end{equation} 
	in which ${\theta_o}, -\pi<\theta_o<=\pi$ is the gradient orientation. Since the angles of the level-lines ${\theta_l}={\theta_o}+\pi/2$ lie in the range of $[-0.5\pi, 1.5\pi]$, they are quantized into eight corresponding directions for searching the candidates of the next edge points, as shown in Fig. \ref{fig_ed}. 
	
\begin{figure}[tbp]
		\centering
		\setlength{\abovecaptionskip}{0.cm}
		\setlength{\belowcaptionskip}{-0.cm}
		\includegraphics[width=0.95\linewidth]{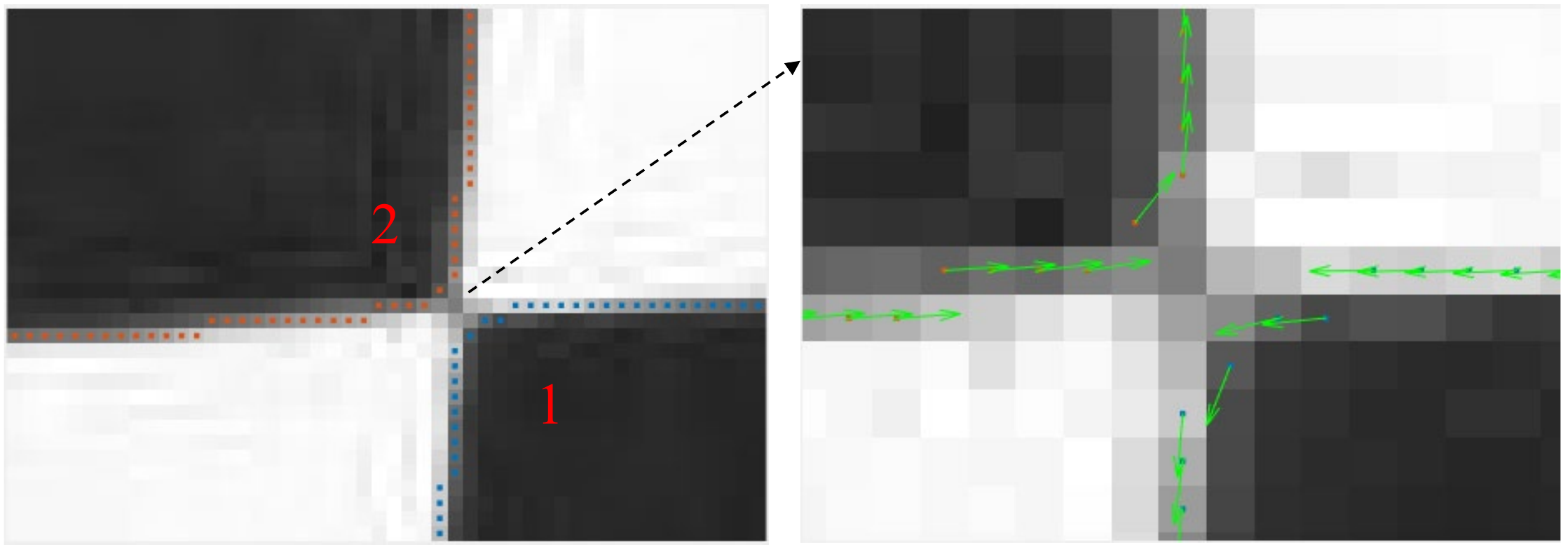}
		\caption{Edges are drawn by the proposed approach, in which the level-lines (green arrows) serve as the drawing guidelines.}
		\label{fig_edges}
	\end{figure}

	The level-lines provide potential directions for accurate edge drawing. Starting from the anchors, for each current edge point, there are three specific searching candidates of the next edge points thanks to the level-lines, instead of the rough searching candidates as in \cite{EDLines} and \cite{ELSED}, significantly when the edge directions are changed. The point with maximal gradient magnitude in searching candidates is selected as the next edge point. The drawing is performed iteratively until all anchors are tracked. Fig. \ref{fig_edges} shows an example of edges drawn by the GEDRLSD algorithm. 
	
	%A video demo regarding the edge drawing is provided in the supplementary material.
	
	\begin{figure}[tbp]
		\centering
		\setlength{\abovecaptionskip}{0.cm}
		\setlength{\belowcaptionskip}{-0.cm}
		\includegraphics[width=1\linewidth]{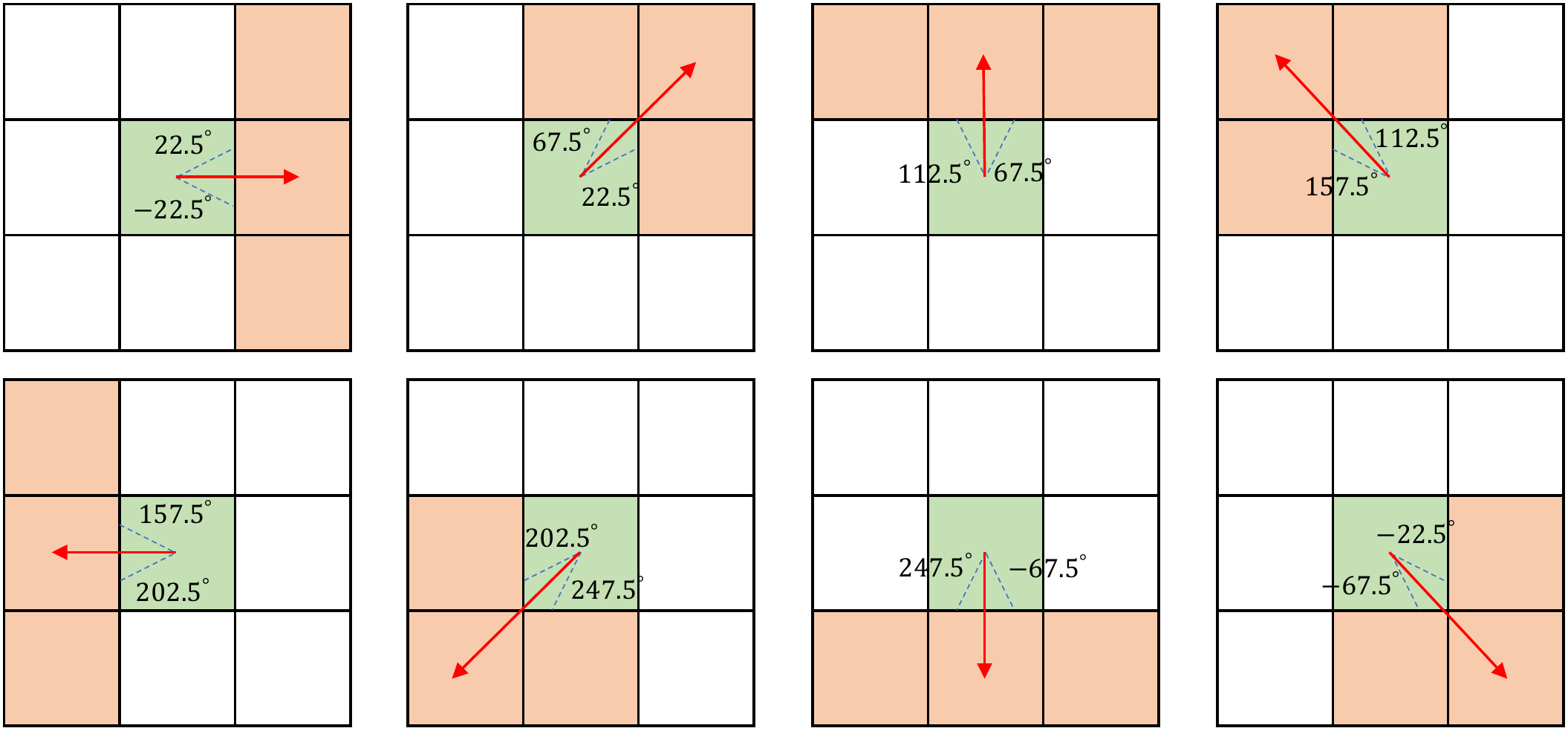}
		\caption{The searching candidates (orange) of the next edge points are based on the level-lines (red arrows, which are quantized within the range of corresponding level-line angles) of the current edge point (green).}
		\label{fig_ed}
	\end{figure}	
	
	\subsection{Edge Refinement}
	\label{sec_III_2}
	As in \cite{CPDA}, if the start  and end points of an edge lie in a threshold range, \textit{i.e.}, 3 pixels, it is declared as the "loop" edge. Otherwise, it is the "line" edge.  Two "line" edges are merged as one if their start and end points lie in a threshold range, \textit{i.e.}, 3 pixels. For "loop" edges, the initial edges may have wrong start and end points due to the ordering of selected anchors. Here, the "loop" edges are reordered according to the corner function introduced in \cite{CPDA}. Specifically, the point with the highest corner function (sharpest corner) is always selected as the start point for "loop" edges, which ensures the single search direction of line segment fitting. 
	
	\subsection{Level-line Leveraged Line Segment Fitting}
	\label{sec_III_3}
	As mentioned above, the line segment in this paper should locate on the edge points with both consistent coordinates and level-lines. Here, the initial line segment is generated by fitting the coordinates of edge points using the least square \cite{ELSED} and validated until the constraint conditions are satisfied. Two validation conditions are set in this paper, \textit{i.e.}, the distance error between the edge coordinates and the fitted lines, and the angle error between the edge level-lines and the fitted lines. Only the ratio of inliers validated by both distances and angles is larger than a threshold $T_{ir}$, the line segment is valid. 
	
	Once the initial line segment is generated and validated, it is grown iteratively using more edge points. Finally, the coordinates and level-lines of all tracked edge points are used to refine the line segment (with parameters of $a$, $b$, and $c$) under the optimization framework according to the formula
	\begin{gather}
		\small
		\setlength{\abovecaptionskip}{3.cm}
		\setlength{\belowcaptionskip}{-3.cm}
		loss  = \min_{a,b,c} \quad \mathcal{F}(\bm{d}) + \rho \times \mathcal{F}(\bm{\theta}),  
		\label{loss}
		\\
		\mathcal{F}(\bm{d})=\left\{
		\begin{aligned}
			1, 					\quad \quad \quad \text{if} \quad \mathcal{F}'(\bm{d}) >= T^d_{v}, \\ 
			\mathcal{F}'(\bm{d})/T^d_{v}, 	\quad \text{if} \quad \mathcal{F}'(\bm{d}) < T^d_{v}, 
		\end{aligned}
		\right.  \\
		\mathcal{F}(\bm{\theta})=\left\{
		\begin{aligned}
			1, 					\quad \quad \quad \text{if} \quad \mathcal{F}'(\bm{\theta}) >= T^a_{v}, \\ 
			\mathcal{F}'(\bm{\theta})/T^a_{v}, \quad	\text{if} \quad \mathcal{F}'(\bm{\theta}) < T^a_{v},
		\end{aligned}
		\right.
	\end{gather} 
	in which $\mathcal{F}'(d) = |a\times x+b\times y+c|/\sqrt{a^2+b^2}$ is normalized regarding distance validation threshold $T^d_{v}$,  and $\mathcal{F}'(\theta)=acos(|-b\times u+a\times v|/(\sqrt{a^2+b^2}\times\sqrt{u^2+v^2}))$ is normalized regarding angle validation threshold $T^a_{v}$. The $(x, y)$ and $(u, v)$ are the coordinate and level-line of an edge point, respectively. The $\rho$ is a weight factor between $\mathcal{F}(\bm{d})$ and $\mathcal{F}(\bm{\theta})$. The line segment endpoints are determined by projecting the first and last inlier edge points onto the fitted line. The refinement process can also be applied to increase the robustness further when the initial line segment is generated.

	\section{Numerical Experiments}
	\label{sec_IV}
	To quantitatively evaluate the proposed GEDRLSD method, various existing SOTA line segment detection methods, including two Hough based methods, \textit{i.e.}, HoughP \cite{RobustDetectionofLinesUsingtheProgressiveProbabilisticHoughTransform} and MCMLSD \cite{MCMLSD}, five local information analysis based methods, \textit{i.e.}, EDLines \cite{EDLines}, ELSED \cite{ELSED}, LSD \cite{LSDaLineSegmentDetector}, FLD \cite{Outdoorplacerecognitioninurbanenvironmentsusingstraightlines}, and Linelet \cite{ANovelLineletBasedRepresentationforLineSegmentDetection}, and four deep learning based methods, \textit{i.e.}, MLSD \cite{TowardsRealtimeandLightweightLineSegmentDetection}, LCNN \cite{End-to-EndWireframeParsing}, HT-LCNN \cite{DeepHoughTransformLinePriors}, and FClip \cite{FullyConvolutionalLineParsing}, are compared based on well-known benchmark datasets.
	
	\subsection{Evaluation Datasets \& Metrics}
	Since line segment detection is a pixel-level task, it is hard to label their ground-truth. Although some datasets \cite{LearningtoParseWireframesinImagesofMan-MadeEnvironments,ANovelLineletBasedRepresentationforLineSegmentDetection} have "ground-truth", their accuracy and correctness are hard to guarantee due to human labeling error and subjectivity. In this paper, inspired by feature point evaluation, the ground-truth free evaluation considering the detection repeatability in different images is adopted to evaluate various methods. The well-known publicly available datasets, \textit{i.e.}, the affine covariant feature dataset \cite{mikolajczyk2005comparison} and HPatches \cite{HPatches} dataset, are selected to perform the quantitative evaluation.
	
	The repeatability defined in \cite{CPDA} is used as the evaluation metric, which can be formulated as $rep = \frac{n_m}{2}\times(\frac{1}{n_r}+\frac{1}{n_t})$, in which $n_m$, $n_r$, and $n_t$ are the numbers of matched line segments, line segments in the reference image, and line segments in the test image, respectively. Two line segments $l_r$ and $l_t$ are matched only if the projected line segment $l'_r$ (according to the Homograph matrix) lies in the neighborhood of $l_t$ with a distance threshold of $T^d_e$ pixels, an angle threshold of $T^a_e$ degrees, and an overlap threshold of $T_o$, as described in \cite{ANovelLineletBasedRepresentationforLineSegmentDetection}. Besides, the line segment matches should be one-to-one, which means that the projected line segments $l'_t$ should also lie in the neighborhood of $l_r$ similarly and that only the mutual closest matched of them are preserved. 
	
	\subsection{Parameters Setting}
	In all experiments, the line segments with lengths smaller than 15 pixels are discarded for all the testing methods to eliminate the effect of short line fragments. In the GEDRLSD algorithm, the normalized gradient magnitude $T^m_g$ is set to 0.2.  When fitting the line segments, the validation parameters, \textit{i.e.}, inlier ratio $T_{ir}$, distance threshold $T^d_v$, and angle threshold $T^a_v$, are set to 0.5, 3 pixels, and 20 degrees, respectively. The weight factor $\rho$ in formula \ref{loss} is 2. For other SOTA methods, the parameters are default values provided by the authors.
	
	\subsection{Results Analysis}
	\begin{figure*}[tbp]
		\centering
		\setlength{\abovecaptionskip}{0.cm}
		\setlength{\belowcaptionskip}{-0.cm}
		\includegraphics[width=1\linewidth]{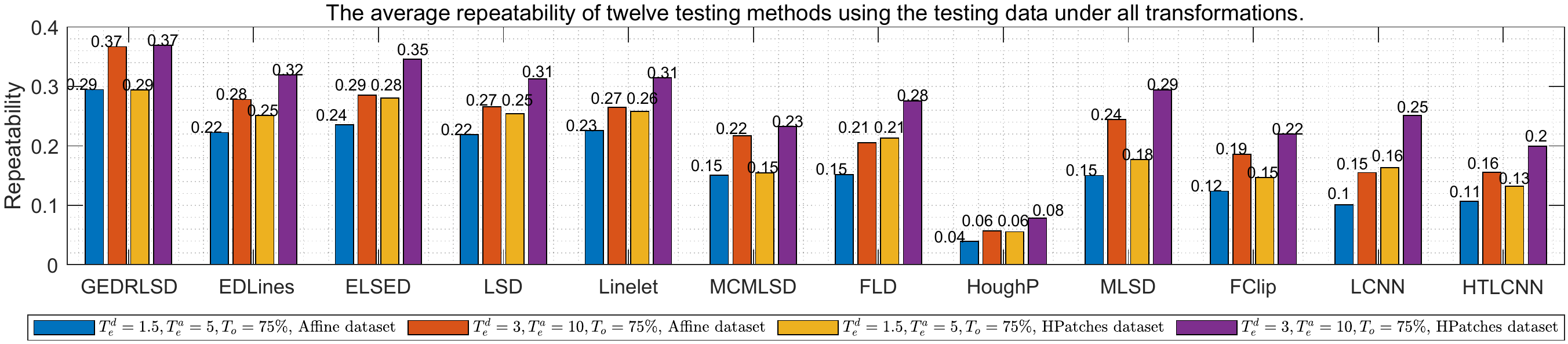}
		\caption{The average repeatability of 12 testing methods based on two evaluation datasets and two evaluation parameter configurations, in which the results of all transformations are considered together.}
		\label{fig_results_1}
		\vspace{-10pt}
	\end{figure*}			
	
	Fig. \ref{fig_line_demo} qualitatively shows an example of line segments detected by the ELSED and proposed GEDRLSD algorithms. Intuitively, the GEDRLSD method detects more complete line segments than the ELSED method. Fig. \ref{fig_results_1} quantitatively shows the average repeatability of twelve methods based on two datasets and two evaluation parameter configurations under all transformations. The results indicate that the proposed GEDRLSD algorithm outperforms other SOTA methods in all the evaluation datasets and configurations. It can effectively and repeatably detect line segments. 
	
	Table \ref{results_table} shows the average repeatability of twelve testing methods under a single transformation. The results indicate that, except for the light transformation, the proposed GEDRLSD algorithm performs best in all the single transformations of two evaluation datasets. The main reason is that image gradients are sensitive to light transformation and insensitive to other abovementioned transformations.

	\begin{table*}[tbp]
		\scriptsize		
		\centering
		\setlength{\abovecaptionskip}{0.cm}
	 	\setlength{\belowcaptionskip}{-0.cm}		
		\caption{The average repeatability of 12 testing methods based on two evaluation datasets and two evaluation parameter configurations, in which the results of the single transformation are considered. The colored values are the top three of performance. }
		\begin{tabular}{|c|ccccccc|ccccccc|}
			\hline
			& \multicolumn{7}{c|}{Evaluation parameter configuration: $T^d_e=1.5, T^a_e=5,   T_o=75\%$}                                                                                                                                                                                                                                                                                                & \multicolumn{7}{c|}{Evaluation parameter configuration: $T^d_e=3, T^a_e=10,   T_o=75\%$}                                                                                                                                                                                                                                                                                                 \\ \cline{2-15} 
			& \multicolumn{5}{c|}{Affine feature   dataset}                                                                                                                                                                                                                     & \multicolumn{2}{c|}{HPatches dataset}                                            & \multicolumn{5}{c|}{Affine feature dataset}                                                                                                                                                                                                                       & \multicolumn{2}{c|}{HPatches dataset}                                            \\ \cline{2-15} 
			\multirow{-3}{*}{} & \multicolumn{1}{c|}{blur}                         & \multicolumn{1}{c|}{view}                         & \multicolumn{1}{c|}{zoom+rotation}                         & \multicolumn{1}{c|}{light}                       & \multicolumn{1}{c|}{JPEG}                         & \multicolumn{1}{c|}{light}                       & {\color[HTML]{000000} view}  & \multicolumn{1}{c|}{blur}                         & \multicolumn{1}{c|}{view}                         & \multicolumn{1}{c|}{zoom+rotation}                         & \multicolumn{1}{c|}{light}                       & \multicolumn{1}{c|}{JPEG}                         & \multicolumn{1}{c|}{light}                       & view                         \\ \hline
			GEDRLSD             & \multicolumn{1}{c|}{{\color[HTML]{FE0000} 0.182}} & \multicolumn{1}{c|}{{\color[HTML]{FE0000} 0.266}} & \multicolumn{1}{c|}{{\color[HTML]{FE0000} 0.165}} & \multicolumn{1}{c|}{{\color[HTML]{34FF34} 0.435}} & \multicolumn{1}{c|}{{\color[HTML]{FE0000} 0.697}} & \multicolumn{1}{c|}{{\color[HTML]{3531FF} 0.281}} & {\color[HTML]{FE0000} 0.307} & \multicolumn{1}{c|}{{\color[HTML]{FE0000} 0.287}} & \multicolumn{1}{c|}{{\color[HTML]{FE0000} 0.368}} & \multicolumn{1}{c|}{{\color[HTML]{FE0000} 0.233}} & \multicolumn{1}{c|}{{\color[HTML]{3531FF} 0.449}}                        & \multicolumn{1}{c|}{{\color[HTML]{FE0000} 0.711}} & \multicolumn{1}{c|}{0.330}                        & {\color[HTML]{FE0000} 0.407} \\ \hline
			EDLines            & \multicolumn{1}{c|}{{\color[HTML]{3531FF} 0.126}} & \multicolumn{1}{c|}{0.209}                        & \multicolumn{1}{c|}{0.121}                        & \multicolumn{1}{c|}{0.403}                        & \multicolumn{1}{c|}{0.464}                        & \multicolumn{1}{c|}{0.258}                        & {\color[HTML]{000000} 0.244} & \multicolumn{1}{c|}{{\color[HTML]{3531FF} 0.182}} & \multicolumn{1}{c|}{{\color[HTML]{3531FF} 0.295}} & \multicolumn{1}{c|}{{\color[HTML]{3531FF} 0.182}} & \multicolumn{1}{c|}{0.428}                        & \multicolumn{1}{c|}{0.477}                        & \multicolumn{1}{c|}{0.310}                        & 0.328                        \\ \hline
			ELSED              & \multicolumn{1}{c|}{0.116}                        & \multicolumn{1}{c|}{{\color[HTML]{34FF34} 0.224}} & \multicolumn{1}{c|}{{\color[HTML]{34FF34} 0.138}} & \multicolumn{1}{c|}{{\color[HTML]{3531FF} 0.406}} & \multicolumn{1}{c|}{{\color[HTML]{34FF34} 0.528}} & \multicolumn{1}{c|}{{\color[HTML]{34FF34} 0.290}} & {\color[HTML]{34FF34} 0.271} & \multicolumn{1}{c|}{0.169}                        & \multicolumn{1}{c|}{{\color[HTML]{34FF34} 0.303}} & \multicolumn{1}{c|}{{\color[HTML]{34FF34} 0.189}} & \multicolumn{1}{c|}{0.423}                        & \multicolumn{1}{c|}{{\color[HTML]{3531FF} 0.536}} & \multicolumn{1}{c|}{{\color[HTML]{3531FF} 0.338}} & {\color[HTML]{34FF34} 0.354} \\ \hline
			LSD                & \multicolumn{1}{c|}{{\color[HTML]{34FF34} 0.127}} & \multicolumn{1}{c|}{{\color[HTML]{3531FF} 0.222}} & \multicolumn{1}{c|}{0.125}                        & \multicolumn{1}{c|}{0.394}                        & \multicolumn{1}{c|}{0.410}                        & \multicolumn{1}{c|}{0.251}                        & {\color[HTML]{3531FF} 0.258} & \multicolumn{1}{c|}{0.174}                        & \multicolumn{1}{c|}{{\color[HTML]{3531FF} 0.295}} & \multicolumn{1}{c|}{{\color[HTML]{3531FF} 0.182}} & \multicolumn{1}{c|}{0.404}                        & \multicolumn{1}{c|}{0.422}                        & \multicolumn{1}{c|}{0.295}                        & {\color[HTML]{3531FF} 0.330} \\ \hline
			Linelet            & \multicolumn{1}{c|}{0.120}                        & \multicolumn{1}{c|}{0.181}                        & \multicolumn{1}{c|}{{\color[HTML]{3531FF} 0.126}} & \multicolumn{1}{c|}{{\color[HTML]{FE0000} 0.473}} & \multicolumn{1}{c|}{{\color[HTML]{3531FF} 0.480}} & \multicolumn{1}{c|}{{\color[HTML]{FE0000} 0.297}} & {\color[HTML]{000000} 0.220} & \multicolumn{1}{c|}{{\color[HTML]{34FF34} 0.185}} & \multicolumn{1}{c|}{0.228}                        & \multicolumn{1}{c|}{0.157}                        & \multicolumn{1}{c|}{{\color[HTML]{34FF34} 0.494}} & \multicolumn{1}{c|}{0.486}                        & \multicolumn{1}{c|}{{\color[HTML]{34FF34} 0.344}} & 0.286                        \\ \hline
			MCMLSD             & \multicolumn{1}{c|}{0.093}                        & \multicolumn{1}{c|}{0.144}                        & \multicolumn{1}{c|}{0.083}                        & \multicolumn{1}{c|}{0.279}                        & \multicolumn{1}{c|}{0.290}                        & \multicolumn{1}{c|}{0.180}                        & {\color[HTML]{000000} 0.130} & \multicolumn{1}{c|}{0.147}                        & \multicolumn{1}{c|}{0.223}                        & \multicolumn{1}{c|}{0.151}                        & \multicolumn{1}{c|}{0.349}                        & \multicolumn{1}{c|}{0.345}                        & \multicolumn{1}{c|}{0.249}                        & 0.216                        \\ \hline
			FLD                & \multicolumn{1}{c|}{0.076}                        & \multicolumn{1}{c|}{0.171}                        & \multicolumn{1}{c|}{0.076}                        & \multicolumn{1}{c|}{0.299}                        & \multicolumn{1}{c|}{0.270}                        & \multicolumn{1}{c|}{0.228}                        & {\color[HTML]{000000} 0.199} & \multicolumn{1}{c|}{0.121}                        & \multicolumn{1}{c|}{0.244}                        & \multicolumn{1}{c|}{0.136}                        & \multicolumn{1}{c|}{0.338}                        & \multicolumn{1}{c|}{0.300}                        & \multicolumn{1}{c|}{0.273}                        & 0.278                        \\ \hline
			HoughP             & \multicolumn{1}{c|}{0.028}                        & \multicolumn{1}{c|}{0.032}                        & \multicolumn{1}{c|}{0.011}                        & \multicolumn{1}{c|}{0.107}                        & \multicolumn{1}{c|}{0.066}                        & \multicolumn{1}{c|}{0.076}                        & {\color[HTML]{000000} 0.036} & \multicolumn{1}{c|}{0.041}                        & \multicolumn{1}{c|}{0.052}                        & \multicolumn{1}{c|}{0.018}                        & \multicolumn{1}{c|}{0.138}                        & \multicolumn{1}{c|}{0.093}                        & \multicolumn{1}{c|}{0.101}                        & 0.056                        \\ \hline
			MLSD               & \multicolumn{1}{c|}{0.085}                        & \multicolumn{1}{c|}{0.064}                        & \multicolumn{1}{c|}{0.046}                        & \multicolumn{1}{c|}{0.363}                        & \multicolumn{1}{c|}{0.449}                        & \multicolumn{1}{c|}{0.265}                        & {\color[HTML]{000000} 0.092} & \multicolumn{1}{c|}{0.173}                        & \multicolumn{1}{c|}{0.129}                        & \multicolumn{1}{c|}{0.116}                        & \multicolumn{1}{c|}{{\color[HTML]{FE0000} 0.555}} & \multicolumn{1}{c|}{{\color[HTML]{34FF34} 0.562}} & \multicolumn{1}{c|}{{\color[HTML]{FE0000} 0.395}} & 0.197                        \\ \hline
			Fclip              & \multicolumn{1}{c|}{0.096}                        & \multicolumn{1}{c|}{0.059}                        & \multicolumn{1}{c|}{0.029}                        & \multicolumn{1}{c|}{0.293}                        & \multicolumn{1}{c|}{0.332}                        & \multicolumn{1}{c|}{0.221}                        & {\color[HTML]{000000} 0.075} & \multicolumn{1}{c|}{0.168}                        & \multicolumn{1}{c|}{0.131}                        & \multicolumn{1}{c|}{0.064}                        & \multicolumn{1}{c|}{0.380}                        & \multicolumn{1}{c|}{0.384}                        & \multicolumn{1}{c|}{0.293}                        & 0.150                        \\ \hline
			LCNN               & \multicolumn{1}{c|}{0.065}                        & \multicolumn{1}{c|}{0.040}                        & \multicolumn{1}{c|}{0.028}                        & \multicolumn{1}{c|}{0.223}                        & \multicolumn{1}{c|}{0.319}                        & \multicolumn{1}{c|}{0.234}                        & {\color[HTML]{000000} 0.096} & \multicolumn{1}{c|}{0.124}                        & \multicolumn{1}{c|}{0.085}                        & \multicolumn{1}{c|}{0.062}                        & \multicolumn{1}{c|}{0.334}                        & \multicolumn{1}{c|}{0.365}                        & \multicolumn{1}{c|}{0.309}                        & 0.195                        \\ \hline
			HTLCNN             & \multicolumn{1}{c|}{0.097}                        & \multicolumn{1}{c|}{0.025}                        & \multicolumn{1}{c|}{0.083}                        & \multicolumn{1}{c|}{0.199}                        & \multicolumn{1}{c|}{0.248}                        & \multicolumn{1}{c|}{0.200}                        & {\color[HTML]{000000} 0.066} & \multicolumn{1}{c|}{0.164}                        & \multicolumn{1}{c|}{0.056}                        & \multicolumn{1}{c|}{0.112}                        & \multicolumn{1}{c|}{0.287}                        & \multicolumn{1}{c|}{0.292}                        & \multicolumn{1}{c|}{0.264}                        & 0.138                        \\ \hline
		\end{tabular}
		\label{results_table}
		\vspace{-6pt}
	\end{table*}	
	
	\subsection{Computation Cost Analysis}
	For edge drawing based line segment detection methods, the primary computation cost comes from the progressive line segment fitting based on the least square using the coordinate of edge points. Here, the additional operation mainly lies in the line segment refinement using both the coordinate of edge points and their level-line information for the GEDRLSD method, which is only performed when the initial line segment is found (optional), and the line segment is stopped to grow. For each line segment, there are at most two times of refinement. The computation cost of the GEDRLSD method is slightly higher than the ELSED and EDLines methods, but their computation costs should be at the same level, much faster than other non-edge drawing based methods.
	
	\subsection{Application in Visual Localization}
	The proposed GEDRLSD algorithm is applied in the long-term visual localization system mentioned in \cite{our_demo}. The road route ”Log 3” of the Ford AV dataset\footnote{https://avdata.ford.com/} is selected to perform the testing. The localization results show that the proposed GEDRLSD algorithm can be successfully applied in a long-term visual localization system, achieving centimeter-level positioning and high orientation accuracy. The details can be found in the published video demo \url{https://github.com/roylin1229/GEDRLSD}.

	\section{Conclusion}
	\label{sec_V}
	This paper proposed the GEDRLSD algorithm by assuming that the line segments should be located on the edge points with consistent coordinates and level-lines. The double consistent constraints make it more robust than the methods based on a single coordinate constraint. The level-lines provide potential directions for edge point tracking and is fused in the line segment refinement. Numerical experiments show that the proposed approach outperforms other SOTA methods while still keeping its efficiency competitive.

	\bibliographystyle{IEEEbib}

\end{document}